\documentclass[5p,authoryear]{elsarticle}
\makeatletter 
\def\ps@pprintTitle{%
 \let\@oddhead\@empty
 \let\@evenhead\@empty
 \let\@evenfoot\@oddfoot} 
\makeatother
\usepackage[utf8]{inputenc} 
\usepackage[T1]{fontenc}
\usepackage[babel=true]{csquotes} 
\usepackage[fleqn]{amsmath} 
\usepackage{amsthm} 
\usepackage{booktabs} 
\usepackage{multirow} 
\usepackage{amssymb} 
\usepackage{float}
\usepackage{hyperref} 
\usepackage[french]{cleveref} 
\usepackage{algorithm}
\usepackage[noend]{algpseudocode}
\usepackage{mathrsfs}
\usepackage{mathtools}

\usepackage{graphicx}
\usepackage{pdflscape,array,booktabs}
\usepackage{threeparttable,booktabs,multirow,lscape}
\usepackage{subfig}
\begin{document}
\begin{frontmatter}
\title{On the Insufficiency of the Large Margins Theory in Explaining the Performance of Ensemble Methods}

\author[add1]{Waldyn Martinez}
\ead{martinwg@miamioh.edu}
\author[add2] {J. Brian Gray}
\address[add1]{Department of Information Systems and Analytics, \\Miami University, Oxford, OH 45056, USA}
\address[add2]{Department of Information Systems, Statistics and Management Science\\ University of Alabama, Tuscaloosa, AL 25205, USA\\}
\cortext[add1]{Corresponding author}

\begin{abstract}
Boosting and other ensemble methods combine a large number of weak classifiers through weighted voting to produce stronger predictive models. To explain the successful performance of boosting algorithms,~\citet{schapire98} showed that AdaBoost is especially effective at increasing the margins of the training data.~\citet{schapire98} also developed an upper bound on the generalization error of any ensemble based on the margins of the training data, from which it was concluded that larger margins should lead to lower generalization error, everything else being equal (sometimes referred to as the ``large margins theory''). Tighter bounds have been derived and have reinforced the large margins theory hypothesis. For instance, \citet {wang2011refined}  suggest that specific margin instances, such as the equilibrium margin, can better summarize the margins distribution. These results have led many researchers to consider direct optimization of the margins to improve ensemble generalization error with mixed results. We show that the large margins theory is not sufficient for explaining the performance of voting classifiers. We do this by illustrating how it is possible to improve upon the margin distribution of an ensemble solution, while keeping the complexity fixed, yet not improve the test set performance.
\end{abstract}


\end{frontmatter}

\section{Introduction}\label{introduction}

Ensemble methods, such as boosting ~\citep{schapire1990strength}, bagging~\citep{breiman1996bagging} and random forests ~\citep{breiman2001random}, along with their variants [see, e.g., stochastic gradient boosting ~\citep{friedman2002stochastic}, rotation forests ~\citep{rodriguez2006rotation}, extreme gradient boosting ~\citep{chen2016xgboost}] create a set of weak classifiers, which are typically decision trees, then combine the predictions from these classifiers in the form of a weighted vote to produce an improved prediction. Empirical evidence suggests that these ensemble methods have lower generalization error than the individual classifiers \citep{drucker1994boosting, dietterich2000ensemble, breiman2001random,  maclin2011popular}. Recently, ensembles have gained further popularity by winning most of the top machine learning competitions in competition platforms such as \textit{Kaggle} ~\citep{hong2014global, puurula2014kaggle, graham2015kaggle, hoch2015ensemble, wang2015large, nielsen2016tree, zou2017youtube} and being applied into a wide range of problems (See e.g., ~\citealt{richardson2000skill, coussement2013customer, king2014ensemble, weng2018macroeconomic, weng2018predicting}). Much theoretical work has been done to explain why ensemble methods are so successful and also why the seemingly complex models often do not overfit \citep{breiman1996bias, schapire98, buja2000special, breiman2000special, mease2008evidence}. In particular, upper bounds on the generalization error of ensemble classifiers based on margins have been developed, which point to margins as a key determinant of ensemble performance. It has been suggested by \citet{schapire98} and \citet{reyzin2006boosting}, among others, that larger margins should result in better ensemble performance, everything else being equal. This has led many researchers to consider directly optimizing the margins to improve ensemble generalization error (see, e.g., ~\citealt{grove98, breiman1999prediction, mason2000improved, ratsch2002maximizing, shen2010boosting, zhou2014large}). Direct optimization of margins has not necessarily yielded successful results, but research is ongoing ~\citep{schapire2013explaining}. Other researchers ~\citep{wang2011refined, wang2012further, gao2013doubt} have provided different generalization error bounds based on functions or particular measures of the margins instances in an attempt to make the theory more complete. In this paper we develop a method for reweighting the weak classifiers of an ensemble in such a way as to increase or maintain all of the margins of a given ensemble solution. According to the large margins theory, this should result in improved performance (lower generalization error) for the combined classifier, if other factors, such as the complexity of the base learners, are held constant. We discover though, that the gains in improved performance are not realized, even when all of the margins of a particular ensemble solution are improved. This leads us to the conclusion that the large margins theory, as currently stated, is insufficient for explaining boosting and ensemble performance. In the next section, we discuss margins, the generalization error bounds based on margins, and the large margins theory. In section 3, we show how to use linear programming techniques to improve or maintain all of the margins of an ensemble and provide empirical evidence to show that this does not necessarily lead to improved ensemble performance. A simulation study is performed and analyzed in section 4. Finally, we discuss our results and their implications for the large margins theory in section 5.\\

\section{Preliminaries}
\label{prelim}
We assume we are given a training sample of data pairs $S=\{(\textbf{x}_1,y_1 ),…,(\textbf{x}_n,y_n )\}$ generated independently and identically distributed (i.i.d) according to an unknown distribution $P_{XY}$ with joint density $p(\textbf{x},y)$, where $Y\in\left\lbrace -1,+1\right\rbrace$ is a binary response variable and $\textbf{x} \in \mathbb{R}^p$. The general goal of learning is to estimate a function $H: X\rightarrow Y$ such that $H$ will correctly classify unseen examples $(\textbf{x},y)$. The function is selected such that the generalization error $R[H]$ (also called the expected risk of the function) is minimized:
\begin{equation}
	\label{eq:1}
	R[H] = E_{\textbf{x},y}g(y,H(\textbf{x}))=\int g(y,H\left( \textbf{x}\right)) dp(\textbf{x},y),
\end{equation}
\noindent where $g(y,H(\textbf{x}))$ is a suitable loss function. For binary classification the loss function $I(y_i \neq H(\textbf{x}))$ is typically used, where $I(y_i \neq H(\textbf{x}))=1$ if $(y_i \neq H(\textbf{x}))$, 0 otherwise. The generalization error cannot be minimized directly because the underlying distribution $P_{XY}$ is unknown. The minimum theoretical value of $R[H]$ is often referred to as Bayes’ minimum risk. $R[H]$ is approximated by the empirical error of the training data set $\hat{R}[H] =\frac{1}{n} \sum g(y,H\left( \textbf{x}\right))$.

We refer to $P[.]$ as probabilities with respect to $P_{XY}$  and $\hat{P}[.]$ as the probability with respect to the empirical distribution over the sample $S$. We assume a set of $T$ classifiers $h_t (\textbf{x}),t = 1,2,...,T$, is created from the space of classifiers $\mathscr{H}$, where $\mathscr{\left|H\right|} < \infty$. The classifiers are usually called base learners, weak learners or weak classifiers, and they are generated from the training data by a base-learning algorithm $\textbf{B}$. Each classifier takes a $p \times 1$ input vector $\textbf{x}$ and produces a prediction $h_t (\textbf{x})\in \{-1,+1\}$  for a binary response variable $Y$. The combined classifier of the prediction is given by the linear function:
\begin{equation}
\label{eq:2}
H(\textbf{x}) = sign \left(\sum_{t=1}^T \alpha_t h_t(\textbf{x})\right)
\end{equation}
\noindent where $sign: \mathbb{R} \rightarrow \left(-1, +1\right)$, such that  $sign(x)=-1$ when $x<0$, $+1$ otherwise, $\alpha_t$ is the weight associated with the $t^{th}$ weak classifier. Without loss of generality, we can assume $0\leq \alpha_t \leq 1$ and $\sum_{t=1}^T \alpha_t =1$. 

The AdaBoost algorithm (ADA) is arguably the best-known boosting algorithm and one of the most used ensemble methods. Unlike earlier boosting methods (\citealt{schapire1990strength, freund1995boosting}), AdaBoost adjusts adaptively to the errors of the weak classifiers (hence the name Adaptive Boosting). AdaBoost is described in Algorithm \ref{alg1}. The task of a boosting is to create a set of weak classifiers and determine their associated weights ${\alpha_1,...,\alpha_T}$ based on the training sample of data $S$, to produce a combined prediction with small generalization error $R[H]$. Many researchers have pointed out the margins of the observations as important characteristics of why the AdaBoost algorithm and most ensemble methods outperform individual classifiers. The $i^{th}$ margins of an observation is given by:
\begin{equation}
\label{eq:3}
m_i = y_i \sum_{t=1}^T \alpha_t h_{it}(\textbf{x}_i).
\end{equation}
A large positive margin of the $i^{th}$ training observation can be viewed as a measure of ``confidence'' in the prediction for the $i^{th}$ training observation \citep{schapire98}. The margin of the $i^{th}$ observation is equal to the difference in the weighted proportion of weak classifiers correctly predicting the $i^{th}$ observation and the weighted proportion of weak classifiers incorrectly predicting the $i^{th}$ observation, so that $-1 \leq m_i \leq 1$. A margin value of $-1$ indicates that all of the weak classifiers' predictions were incorrect, while a margin value of $+1$ indicates all of the weak classifiers correctly predicted the observation.

\begin{algorithm}
\caption{AdaBoost (ADA)}\label{alg1}
\begin{algorithmic}[1]
    \Procedure{ADA}{$S,T$}
		\State  $D_i^{(1)} \gets \frac{1}{n}$
		\State \textbf{do for} $t = 1,...,T$
		\State \indent (a) $h_t(\textbf{x}_i)) \gets h(\{S, D^{(t)}\})$
		\State \indent (b) $\epsilon_t \gets \sum_{i=1}^{n} D_i^{(t)}I \left(y_i \neq h_t(\textbf{x}_i)\right)$
		\State \indent (c) break if  $\epsilon_t = 0$ or $\epsilon_t \geq \frac{1}{2}$
		\State \indent (d) $\alpha_t \gets \frac{1}{2} \ln \left(\frac{1-\epsilon_t}{\epsilon_t}\right)$ 
		\State \indent (e) $D_i^{(t+1)} \gets \frac{ D_i^{(t)}\exp\{-\alpha_ty_ih_t(\textbf{x}_i)\}}{Z_t}$
		\State \textbf{end for}
    \State $H{\textbf{(x)}}= sign  \left(\sum_{t=1}^T\alpha_th_t(\textbf{x})\right)$

\EndProcedure
\State \textbf{return} $H{\textbf{(x)}}$
\end{algorithmic}
\end{algorithm}

 \noindent The development of boosting algorithms was based on the PAC learning theory \citep{valiant1984theory}. It has been shown that AdaBoost is a PAC (strong) learner. as defined below:
\\

\noindent
{\bf Definition 1} {\it \citep{valiant1984theory}. Let $\mathscr{F}$ be a class of concepts. For every distribution $P_{XY}$, all concepts $f \in \mathscr{F}$  and all $\epsilon \in (0,1/2)$, $\delta \in(0,1/2)$, a strong PAC learner has the property that with probability at least $1-\delta$ the base learning algorithm $\textbf{B}$ outputs a hypothesis $h$ with $P\left(h(\textbf{x})\neq f(\textbf{x})\right) \leq \epsilon$. $\textbf{B}$ must run in polynomial time in $1/\epsilon$, and $1/\delta$ using only a polynomial (in $1/\epsilon$ and $1/\delta$) number of examples.}
\\

\noindent A weak learner drops the strong accuracy requirement, that is, to output a hypothesis $h$ with $P\left(h(\textbf{x})\neq f(\textbf{x})\right) \leq \epsilon$ with a probability of at least $1-\delta$, and replaces it with the requirement that the algorithm outputs a hypothesis that performs better than random guessing.

\section{Generalization Error Bounds Based on Margins}
\label{bounds}
Boosting is based on the question posed by \cite{kearns1994cryptographic} about whether weak and strong learning are equivalent for efficient learning algorithms. In the development of the first algorithm that adaptively boosted a weak learner into a stronger performing algorithm, \cite{freund97} answered this question and also gave a bound on the generalization error of boosting in terms of the number of boosting rounds $T$, the VC-dimension $d$ (a measure of complexity) and the training set error rate $\hat{P}\left[H(\textbf{x})\neq y \right]$. (See \citealt{freund97} for the derivation of the bound.) They showed the generalization error, with high probability, was bounded by
\begin{equation}
\label{eq:4}
R[H] \leq \hat{P}\left[H(\textbf{x})\neq y \right] + O\left( \sqrt{\frac{Td}{n}}\right).
\end{equation}
The bound in (\ref{eq:4}) suggests the generalization error would degrade as the number of rounds of boosting $T$ increased. This in fact happens, especially in cases where there is noise, (see, e.g., \citealt{grove98, dietterich2000ensemble, long2010random, martinez2016noise}),  but under certain conditions boosting methods have shown not to overfit even when thousands of rounds are run (see, e.g., \citealt {breiman1996bias, drucker1996boosting, quinlan1996bagging, grove98, opitz1999popular, buja2000special, lugosi2004bayes}). To better explain the effectiveness of AdaBoost and other ensembles, \cite{schapire98} developed an upper bound on the generalization error of any ensemble method, based on the margins of the training data, from which it was concluded that larger margins should lead to a lower generalization error of the ensemble, everything else being equal. This bound does not depend on the number of rounds $T$ and is summarized in theorem 1.
\\

\noindent
{\bf Bound 1}{\it \citep{schapire98}. Assuming that the base-classifier space $\mathscr{H}$ is finite, and for any  $\delta> 0$ and $\theta>0$, then with probability at least $1-\delta$ over the training set $S$ with size $n$, every voting classifier $H$ satisfies the following bound:}
\begin{equation}
\label{eq:5}
R[H] \leq \hat{P}\left[m(\textbf{x},y) \leq \theta) \right] + O\left( \frac{1}{\sqrt{n}}\sqrt{\frac{\ln{n}\ln{\mathscr{\left|H\right|}}} {\theta^2}+\ln{\frac{1}{\delta}}}{}\right),
\end{equation}
\noindent where the term $\hat{P}\left[m(\textbf{x},y) \leq \theta) \right]$ is the proportion of training margins less than an arbitrary value $\theta >0$, and $\mathscr{\left|H\right|}$ refers to the cardinality of the finite hypothesis space. When the hypothesis space is infinite, the term $\ln{n}\ln{\mathscr{\left|H\right|}}$ is replaced by $d \log_2 (d/n)$, where $d$ is the VC-dimension of the space of all possible weak classifiers. \cite{schapire98} use this bound to provide an explanation for the superior performance of AdaBoost, which they show is highly effective at increasing the margins.

\begin{figure}[h]
	\centering
	\includegraphics[width=9cm,height=6cm]{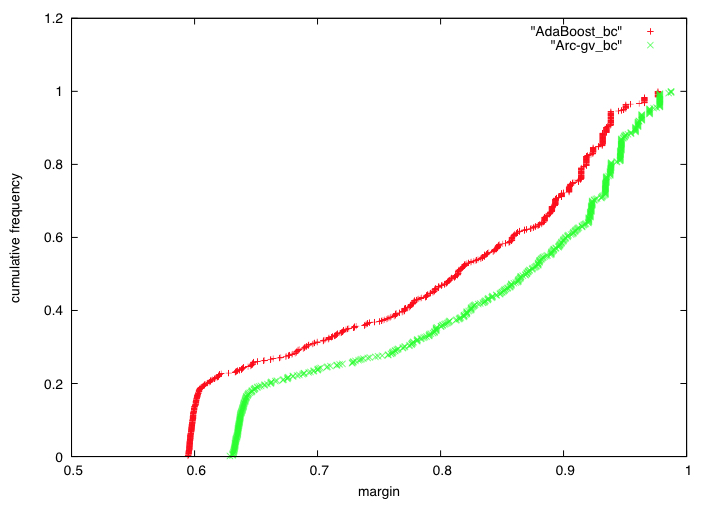}
	\caption{Reprinted from \cite{reyzin2006boosting} with permission (permission will be obtained prior to publication). Cumulative margins for AdaBoost and arc-gv for the breast cancer data set after 500 iterations.}
	\label{fig1} 
\end{figure}

Even though the upper bound in (\ref{eq:5}) is quite loose in most practical situations \citep{breiman1999prediction}, researchers have used it to conclude that higher margins should lead to a lower generalization error rate, everything else being equal. For example, \cite{reyzin2006boosting} state that ``the margins explanation basically says that when all other factors are equal, higher margins result in lower error.'' This interpretation of the upper bound in (\ref{eq:5}) has been referred to as the ``large margins theory"  in the boosting literature. (see, e.g., \citealt{schapire98, grove98, mason2000improved, wang2011refined, shen2010boosting, wang2011refined, cid2012three, wang2012further, gao2013doubt, martinez2014role, liu2015boosting, martinez2019Current}; and \citealt{zhang2016optimal} for further insight into the importance of large margins in ensemble performance.)
Unfortunately, much of the boosting literature is replete with phrases like ``higher margins'', ``large margins for most of the examples'', ``maximizing the number of training examples with large margins'', and ``maximizing the margins'' without specific, operational definitions of these terms (see, e.g.,  \citealt{schapire98}). \cite{grove98} and others, initially defined ``maximizing the margin'' as maximizing the minimum margin. Their linear programming approach, LP-Boost, is designed to maximize the minimum margin by optimizing the weights associated with the weak learners. \cite{breiman1999prediction}, for instance, developed a new boosting algorithm, called arc-gv, which he proved maximizes the minimum margin. Instead of using direct optimization of the margin, \cite{breiman1999prediction} modified the weights of the individual classifiers to be updated to
\begin{equation}
\label{eq:6}
\alpha^*_t= \frac{1}{2} \ln\left(\frac{1+\gamma_t}{1-\gamma_t}\right)-\frac{1}{2}  \ln\left(\frac{1+\rho_t}{1-\rho_t}\right),
\end{equation}

\noindent where $\gamma_t=\sum_{i=1}^n D_i^t y_i h_t (\textbf{x}_i)$ is called the edge of $h_t$, and $\rho_t=\min\left(y_i \sum_{t=1}^T \alpha^*_t h_t (\textbf{x}_i)\right)$. Note that in Algorithm \ref{alg1}, the unnormalized $\alpha_t$ weights can be expressed as $\alpha_t= \frac{1}{2} \ln\left(\frac{1+\gamma_t}{1-\gamma_t}\right)$, so that (\ref{eq:6}) updates the original $\alpha_t$ with information on the minimum margin over all training examples. Breiman also presented a bound on the generalization error of any voting classifier based on the minimum margin that was tighter than that presented by \cite{schapire98}. \cite{breiman1999prediction} found in his experiments that arc-gv not only produced larger minimum margins over all training examples, but it also produced a better margin distribution than AdaBoost (see Figure \ref{fig1}), however, he discovered that the test set error performance of arc-gv was typically worse than for AdaBoost. \cite{breiman1999prediction}  noted that the upper bounds on the generalization error of ensembles proposed by \cite{schapire98} and \cite{breiman1999prediction} are ``greater than one in all practical cases, leaving ample room for other factors to influence the true generalization error'' and concluded that this “casts doubt on the ability of the loose VC-type bounds to uncover the mechanism leading to low generalization error.'' \cite{breiman1999prediction} concluded that the large margins theory was incorrect based on his findings.

\begin{figure}[h]
	\centering
	\includegraphics[width=9cm,height=6cm]{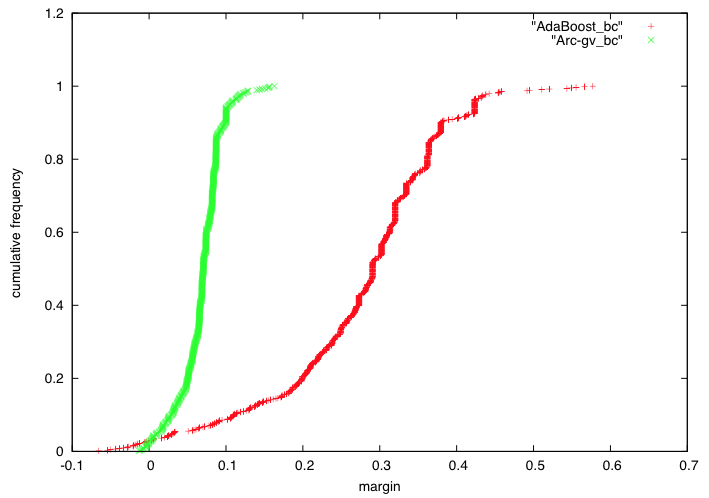}
	\caption{Reprinted from \cite{reyzin2006boosting} with permission (permission will be obtained prior to publication). Cumulative margins for AdaBoost and arc-gv for the breast cancer data set after 100 iterations using decision stumps.}
	\label{fig2} 
\end{figure}
\noindent \cite{reyzin2006boosting} replicated the analysis presented in \cite{breiman1999prediction}, but pointed out that the trees (weak classifiers) found by arc-gv were deeper on average than the trees generated by AdaBoost, even though the number of terminal nodes was kept the same. They concluded that this increased complexity could have led to overfitting by arc-gv, and hence the worse test set performance. More importantly, \cite{reyzin2006boosting} suggested that the increased complexity in the trees generated by arc-gv violated the assumption of everything else kept constant, discrediting Breiman's evidence against the large margins theory. When \cite{reyzin2006boosting} limited both arc-gv and AdaBoost to decision stumps, it was found that while arc-gv did have a larger minimum margin than AdaBoost, it did not produce uniformly higher margins (see Figure \ref{fig2}), furthering their conclusion that the large margins theory was still intact. \cite{reyzin2006boosting} also concluded that maximizing the minimum margin does not necessarily result in improved generalization error.

After the less-than-satisfactory results from maximizing the minimum margin, many authors have supported the large margins theory and have proposed optimizing other functions of the margin distribution. \cite{reyzin2006boosting} suggested maximizing the average or the median margin.  \cite{mason2000improved} proposed the DOOM (Direct Optimization of Margins) algorithm, which optimizes the average of a cost function of the margins. DOOM outperformed AdaBoost in most of the examples considered. They also showed that the size of the minimum margin was not a critical factor in generalization error. \cite{shen2010boosting} suggested an algorithm named MD-Boost (Margin Distribution Boosting), which attempts to maximize the average margin while minimizing the variance of the margin distribution. \cite{shen2010boosting} provide evidence that MD-Boost outperforms AdaBoost in many of the UCI Repository data sets, however it is worth mentioning that the margins produced by MD-Boost are not uniformly larger than those of AdaBoost and consequently the MD-Boost margin distributions are not uniformly larger than those of AdaBoost. Other researchers have summarized the margin distribution of an ensemble on a single metric. For example, \cite{wang2011refined} provided an upper bound for the generalization error of ensemble methods based on a new margin notion referred to as the equilibrium margin (EMargin). To better understand the bound, we introduce the Bernoulli Kullback-Leibler function $D(q\left|\right|p)$ defined as
\begin{equation}
\label{eq:8}
D(q\left|\right|p)\leq q \ln \frac{q}{p}+(1-q)  \ln \frac{1-q}{1-p}, 0\leq p,q\leq 1. 
\end{equation}                             

\noindent $D(q\left|\right|p)$ is a monotone increasing function for a fixed $q$ and $q \leq p < 1$. We can also see that $D(q\left|\right|p)=0$  when $p = q$ and $D(q\left|\right|p) \rightarrow \infty$  as $p\rightarrow 1$. The bound is presented in Theorem 3.
\\

\noindent
{\bf Theorem 3} {\it \citep{wang2011refined}. Assuming that the base-classifier space $\mathscr{\left|H\right|}$ is finite, and for any  $\delta> 0$ and $\theta >0$, then with probability at least $1-\delta$ over the training set $S$ with size $n$, every voting classifier $H$ satisfies the following bound:}
\\
\begin{equation}
\label{eq:9}
R[H]\leq \frac{\ln{\mathscr{\left|H\right|}}}{n}+\inf_{q\in \left(0, \frac{1}{n}, \frac{2}{n},...,1\right)}D^{-1} \left(q;u \left[\hat{\theta}(q)\right]\right),
\end{equation}

\noindent where 
\begin{equation}
\label{eq:10}
u\left[\hat{\theta}(q)\right]= \frac{1}{n} \left( \frac{8\ln\mathscr{\left|H\right|}}{\hat{\theta}^2(q)} \ln \left(\frac{2n^2}{\ln \mathscr{\left|H\right|}}\right)+\ln{\mathscr{\left|H\right|}}+\ln \frac{1}{\delta} \right)
\end{equation}
 and 
\begin{equation}
\label{eq:11}
\hat{\theta}(q)= \sup_{\theta \in \left( \sqrt{\frac{8}{\mathscr{ \left| H \right|}}},1\right]}  \hat{P}\left[m(\textbf{x},y) \leq \theta) \right] \leq q.
\end{equation}
The optimal value of $q$ in (\ref{eq:9}) defined as $q^*$ evaluated at $\hat \theta(q^*)$ is called the equilibrium margin (EMargin), while $q^*$  is called the EMargin error. The upper bound developed by \cite{wang2011refined} is an explanation of the performance of ensemble methods based on a single characteristic called the EMargin, instead of the margin mean and variance. \cite{wang2011refined, wang2012further} provide evidence that ensembles with higher EMargins, $\hat \theta(q^*)$, or lower in EMargin errors $q^*$ would result in a better generalization performance, everything else being equal. In particular they show that AdaBoost's EMargins are higher in most cases than arc-gv's \citep{breiman1999prediction}, which explains why despite arc-gv's larger minimum margin, it has worse performance. \cite{wang2011refined} give further validation to the conjecture that the whole margin distribution is more important than the minimum margin, and that the margin distribution can be better summarized in the single Emargin characteristic as opposed to the sample moments. Tighter bounds under the large margins theory are commonplace in the recent research literature and beyond the scope of this research. Among the most notable contributions is the bound provided by \cite{gao2013doubt}, which suggests that both the bounds in Theorem 2 and Theorem 3 are special cases of the $k^{th}$margin bound. \cite{gao2013doubt} found that the $k^{th}$ margin bound was tighter than the bounds analyzed in Theorem 2 and Theorem 3. \cite{gao2013doubt} further support the idea that the whole margin distribution is the main driver to ensemble method performance and also give further validation to the Emargin notion. 

In the next section, we show how it is possible to improve over the margins of an ensemble solution or its EMargin using the same set of weak learners generated by the original ensemble and still not achieve a better generalization performance. It is important to note that this paper does not attempt to prove the large margins theory wrong, but to shed light on the fact that there may be other unknown factors involved in the explanation of ensemble performance.

\section{Improving the Margins of an Ensemble}
\label{proposed}
We describe here a method, based on linear programming, for improving or at least maintaining all the margins of any ensemble. We do this by simply updating the weights given to the weak classifiers. Once the entire margin distribution is improved, or at least maintained, any summary location measure of the margins distribution (including the minimum, median, mean, and percentiles) will also be improved or maintained. Thus, the proposed method achieves “higher margins” regardless of the definition. Furthermore, we fix the ensemble complexity by using the same set of trees provided by the ensemble, and also by forcing the trees to grow to a fixed depth and number of terminal nodes $k$. We should mention that our simulations results hold regardless of the selected value of $k$, and that the choices of $k$ here are only meant to provide the reader with a better intuition on how the results hold. The choice of a fixed $k$ for a given depth ensures the complexity is unchanged, and allows us to compute the cardinality of the hypothesis space $\mathscr{\left|H\right|}$  when comparing across methods \citep{reyzin2006boosting}, but we also provide evidence on how our results hold for unpruned fully-grown trees. Furthermore, by using linear programming, the proposed methods provide an improvement on the overall computational burden of the resulting ensemble solution, as many of the weights are zeroed out by the optimization.


\subsection{Maximizing the Minimum Margin Improvement}

We assume that we are given an ensemble solution, i.e., a set of weak classifiers $h_t(\textbf{x}),t = 1,2,...,T$, and a set of weights $\alpha_t,t = 1,2,...,T$, where $0\leq \alpha_t \leq 1$ and $\sum_{t=1}^T \alpha_t = 1$, associated with the weak classifiers. Note that for the training sample $\{( \textbf{x}_i, y_i ), i = 1,2, ...,n \}$ used to produce the ensemble solution, the values of the weak classifier predictions and the weights are fixed. We will let $h_{t}(\textbf{x}_i)=\pm 1$  denote the prediction of the $t^{th}$  weak classifier for the $i^{th}$ observation in the training data when given the covariate vector $\textbf{x}_i$. Our goal in the first improvement algorithm, MMI, is optimize the weights associated with the weak classifiers such that we maximize the minimum improvement to the ensemble margins. The linear program (LP) used to achieve this solution is given in Algorithm \ref{MMI}.

\begin{algorithm}
	\caption{MMI}
	\label{MMI}
	\[ \begin{array}{rl}
	\max & \xi \\
	\mbox{s.t.} & y_i \sum{_{t=1}^T \alpha_t' h_{t}(\textbf{x}_i) \geq m_i + \xi}, i = 1,2,..., n\\
	& \sum{_{t=1}^T \alpha_t' = 1}\\
	& \alpha_t' \geq 0, t =1,2,..., T\\
	& \xi \geq 0,
	\end{array}
	\]
\end{algorithm} 

\begin{figure*} 
	 \includegraphics[width=\textwidth]{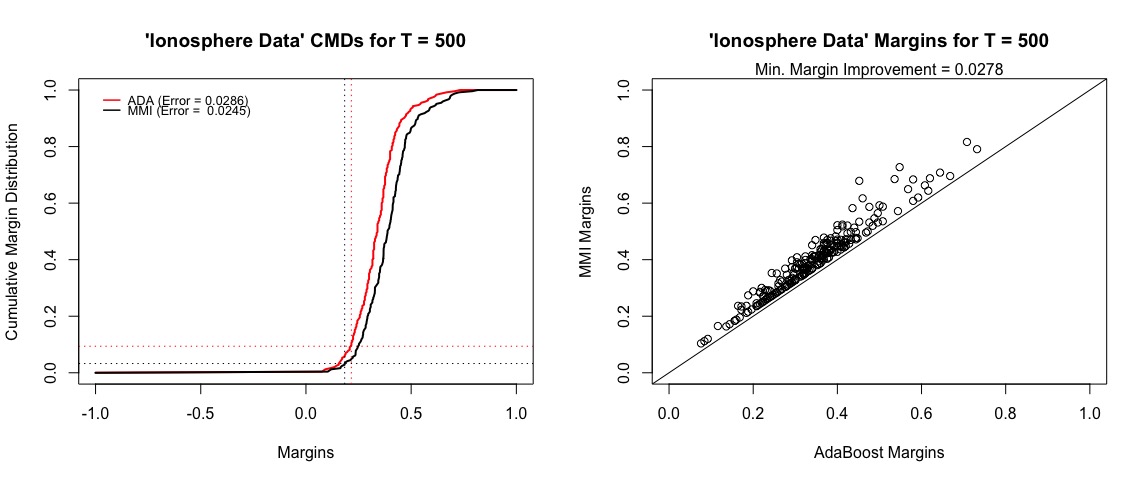}
	\caption{Comparison of AdaBoost and the MMI algorithm for the Ionosphere Data using 750 4-node (depth = 2) trees. The left panel compares the cumulative margin distributions (CMD) for AdaBoost and the MMI algorithm. The dotted lines indicate the Emargins (x-axis) and the Emargin Errors (y-axis). The right panel is a scatter plot of the AdaBoost margins and MMI margins. The line in the graph indicates equality of the margins.}
	\label{fig:3}
\end{figure*}

\noindent In Algorithm \ref{MMI} $m_i = y_i \sum_{t=1}^T \alpha_t h_{t}(\textbf{x}_i)$ is the original ensemble margin for the $i^{th}$ observation and $\alpha_t' , t =1,2,..., T$, are the new weights for the weak learners, to be determined by solving the LP. The linear optimization constraint $y_i \sum{_{t=1}^T \alpha_t' h_{it}(\textbf{x}_i) \geq m_i + \xi}$ for $\xi \geq 0$ guarantees that the choice of $\alpha_t'$ will produce new margins at least as large as the margins generated by ensemble. To illustrate the performance of the MMI algorithm and compare it to original ensemble solution, we fit an AdaBoost model using 750 weak classifiers (depth = 2, 4-node trees) to the Ionosphere data set, then obtain the updated weights and margins from Algorithm \ref{MMI}. The cumulative margin distributions (CMD) of AdaBoost and MMI are plotted in the left panel of Figure \ref{fig:3}. Since the margins generated by the MMI algorithm are uniformly larger, the MMI CMD is to the right of the AdaBoost CMD at all points. We also show a scatter plot of the margins generated by AdaBoost and those for the MMI algorithm in the right panel of Figure \ref{fig:3}. For this particular example, all of the margins were improved (the minimum improvement to the margins was 0.0333), the test set error rate for the MMI algorithm was 0.0566, which is better than the original AdaBoost solution (0.0660). Out of the $t=750$ weak learners combined, the LP solution of the MMI algorithm contained only 90 nonzero $\alpha_t'$ values, reducing the overall size of the ensemble (only 12\% of the decision trees are used in the resulting ensemble solution). In addition to that, the Emargins $\hat \theta(q^*)$ and EMargin Errors $q^*$ for both solutions are $\hat \theta_{\text{ADA}}(q^*) = 0.2500$, with $q_{\text{ADA}}^* = 0.2367$, while $\hat \theta_{\text{MMI}}(q^*) = 0.2888$ and $q_{\text{MMI}}^* =0.1918$. The results in Theorem 3 suggest a voting classifier with larger Emargin and a smaller Emargin error should have a better generalization error holding everything else constant. In this particular example the EMargin for the MMI optimized solution is larger, and the EMargin error is lower supporting the findings in both \cite{schapire98} and \cite{wang2011refined}.

\begin{figure*}[h]
	\centering
	\includegraphics[width=\textwidth]{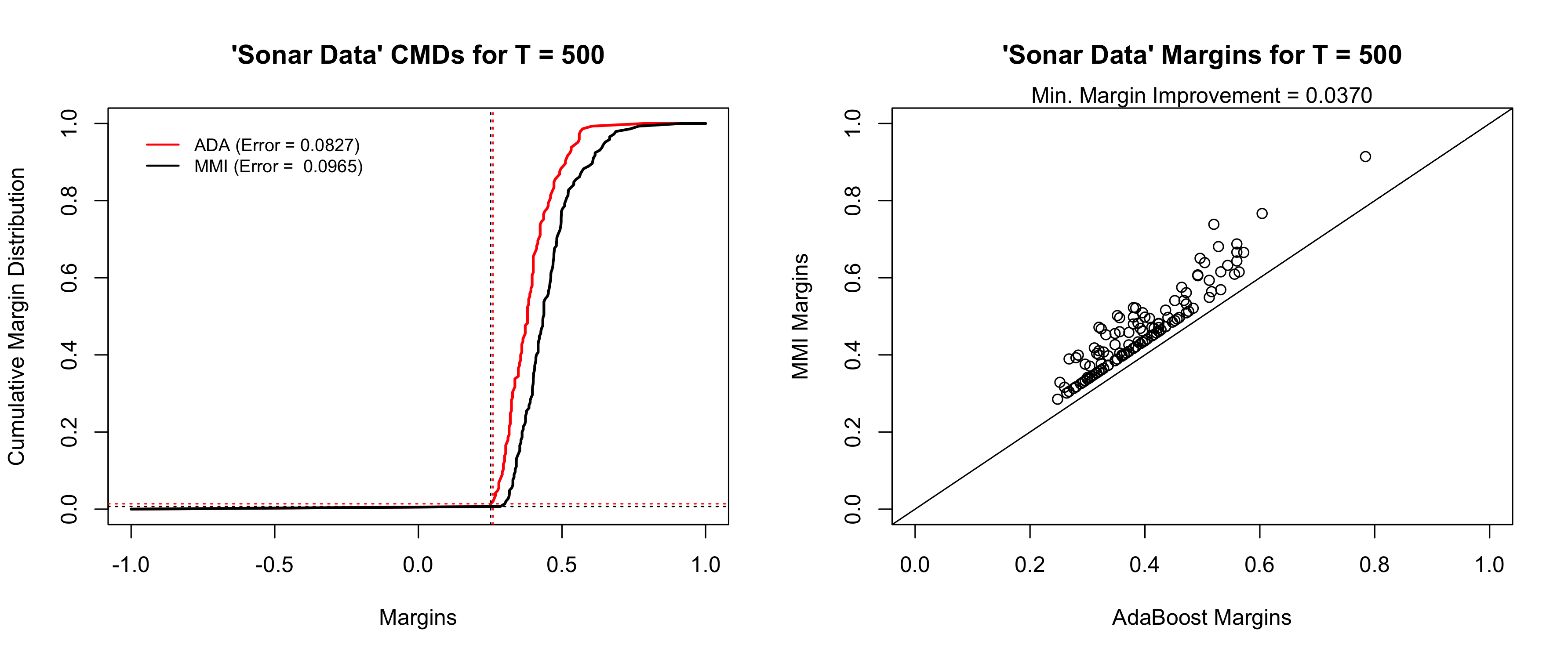}
	\caption{Comparison of AdaBoost and the MMI algorithm for the Ionosphere Data using 500 16-node (depth = 4) trees. The left panel compares the cumulative margin distributions (CMD) for AdaBoost and the MMI algorithm. The dotted lines indicate the Emargins (x-axis) and the Emargin Errors (y-axis). The right panel is a scatter plot of the AdaBoost margins and MMI margins. The line in the graph indicates equality of the margins.}
	\label{fig:4} 
\end{figure*}

Unlike Breiman's arc-gv algorithm \citep{breiman1999prediction}, the MMI algorithm holds the complexity of the weak learners fixed by utilizing the same weak classifiers created by AdaBoost. We further guard against difference in complexities by restricting the analysis to a fixed depth with fixed number of terminal nodes (i.e., $k=4$ and depth = 2 in this example). The performance of the MMI algorithm in this particular example is ideal, considering it uses significantly fewer trees than the original AdaBoost solution, while improving upon the test set error rate. A generalization of this performance under different settings (i.e., different data sets, ensemble sizes, etc.) would strongly support the large margins theory and help us better design ensembles. Notwithstanding, we find that a better performance for the MMI algorithm is not always realized, and, in fact, the opposite is most often true. Consider, for example, Figure \ref{fig:4}, which shows the performance of AdaBoost versus the MMI algorithm for the Australian Credit data set using 500 decision trees forced to have 16 terminal nodes ($k=16$, depth = 4). We note that all margin instances are higher than those of the original AdaBoost solution (the minimum margin improvement was 0.0302), yet the test set error rate for AdaBoost was better (0.1106) than that of the MMI algorithm (0.1442). The percentage of trees used by the MMI algorithm was also a fraction of the original as in the previous example, with only 94 out of the 500 trees with nonzero weights (just about 18.8\% of the original weak classifiers). We also find this example contradicts the premise that a larger EMargin with lower EMargin Error should result in lower generalization performance contradicting the findings in \cite{wang2011refined}. The EMargin and EMargin error for the MMI algorithm are: $\hat \theta_{\text{ADA}}(q^*) = 0.2360$ with $q_{\text{ADA}}^* = 0.5228$, while $\hat \theta_{\text{MMI}}(q^*) = 0.2834$ and $q_{\text{MMI}}^* = 0.4066$. In other words, the EMargin for the MMI algorithm is higher with lower EMargin error.

\begin{figure*}[h]
	\centering
	\includegraphics[width=20cm]{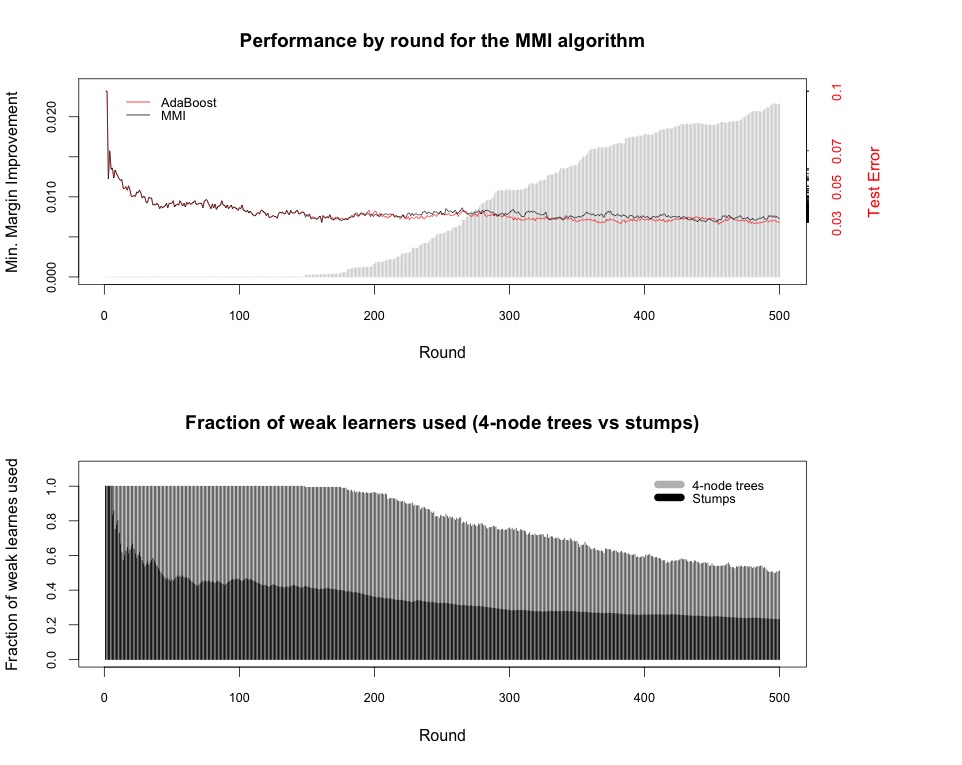}
	\caption{(Top Plot) By round comparison of AdaBoost and MMI algorithm for the Splice data set using decision trees with 4 terminal nodes (depth = 2). Lines indicate error rates and the density plot indicates minimum margin improvements. (Bottom Plot) Fraction of weak learners used for 4 terminal nodes (depth = 2) trees vs stumps (depth = 1).}
	\label{fig:5} 
\end{figure*}

In the previous examples we have found contradicting information regarding ensemble performance. We have empirically shown how it is possible to improve or maintain all of the margins in an ensemble solution by adjusting the weak classifier weights, yet have the same or worse test set performance. We have done this while keeping the complexity fixed. The lack of improvement in the test error rate contradicts the large margins theory, regardless of how ``higher margins" is defined. Nevertheless, an example does not necessarily prove the theory is incorrect or incomplete. There are many questions unanswered including whether a better performance is only guaranteed in specific circumstances such as shallower trees, fewer iterations, or larger sample sizes. In the next section, we use simulation to find answers to these questions. We attempt to answer the question posed by \cite{wang2011refined}, which stated ``can we find a strategy that optimizes the margin distribution? If such an algorithm exists, it would be a good test of our theory to see whether it has better performance than AdaBoost as we predict."


\section{Experiments and Simulations}
\label{sims}

To further understand how the performance of the MMI algorithm changes as the number of weak learners increases, we plot the test set error rate for both the original AdaBoost solution and the MMI algorithm for $t=1,...,500$ on the Splice data set (see section 5 for description). The upper panel in Figure \ref{fig:5} shows that as the number of weak learners increases, the minimum margin improvement to the AdaBoost solution also increases. However the MMI algorithm does not outperform AdaBoost as the large margins theory suggest, in fact more often than not AdaBoost has a lower test set error rate. The lower panel on Figure \ref{fig:5} shows how the choice of tree depth affects the fraction of weak learners used given the number of weak learners combined. As the number of trees combined increases, the fraction of trees utilized decreases at a faster rate in stumps (depth = 1) when compared to 4-node (depth = 2) decision trees. This may be due to the fact that the LP solution more easily finds similarity in the stumps, as opposed to the 4-node (depth = 2) trees. 

\begin{figure}
\begin{center}
  \subfloat{
	\begin{minipage}[c][1\width]{
	   0.33\textwidth}
	   \centering
	   \includegraphics[width=6.2in, height = 2.4in]{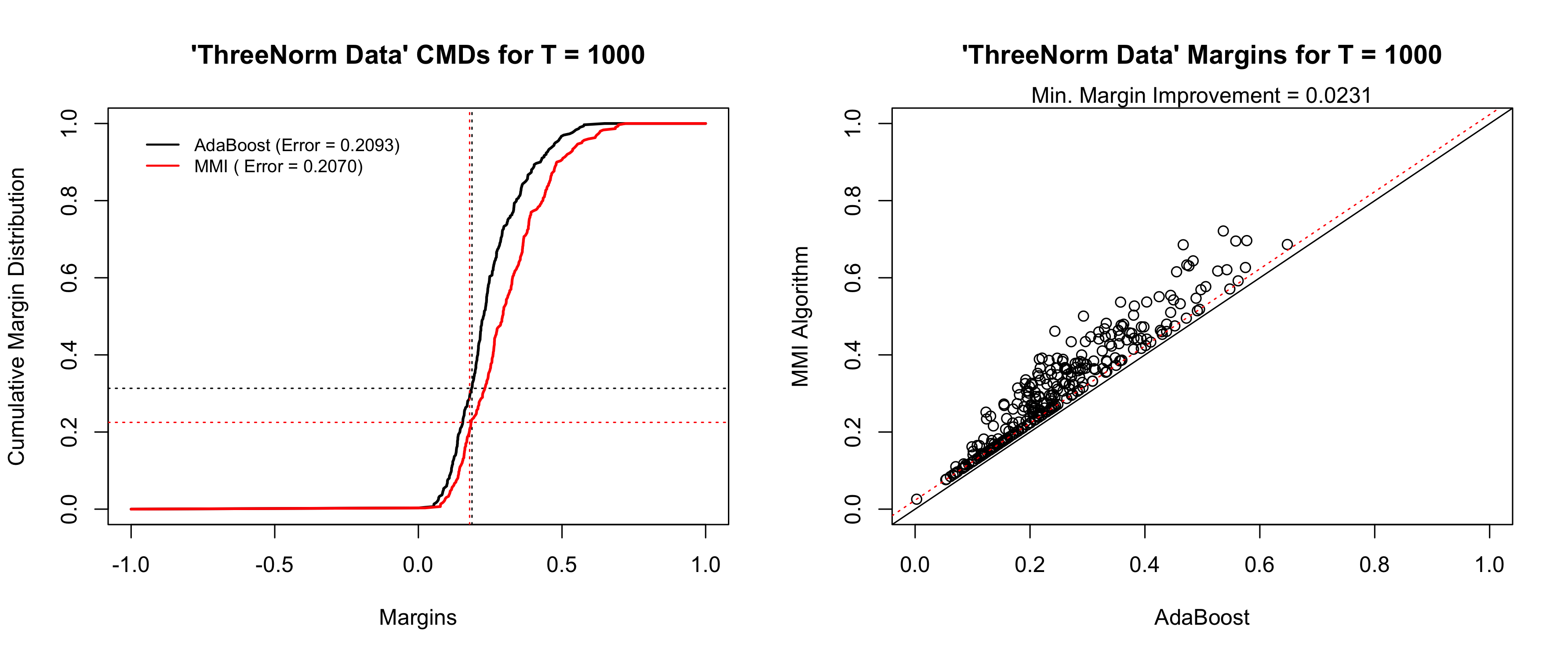}
	\end{minipage}}
 \vfill 	
  \subfloat{
	\begin{minipage}[c][1\width]{
	   0.33\textwidth}
	   \centering
	   \includegraphics[width=6.2in, height = 2.4in]{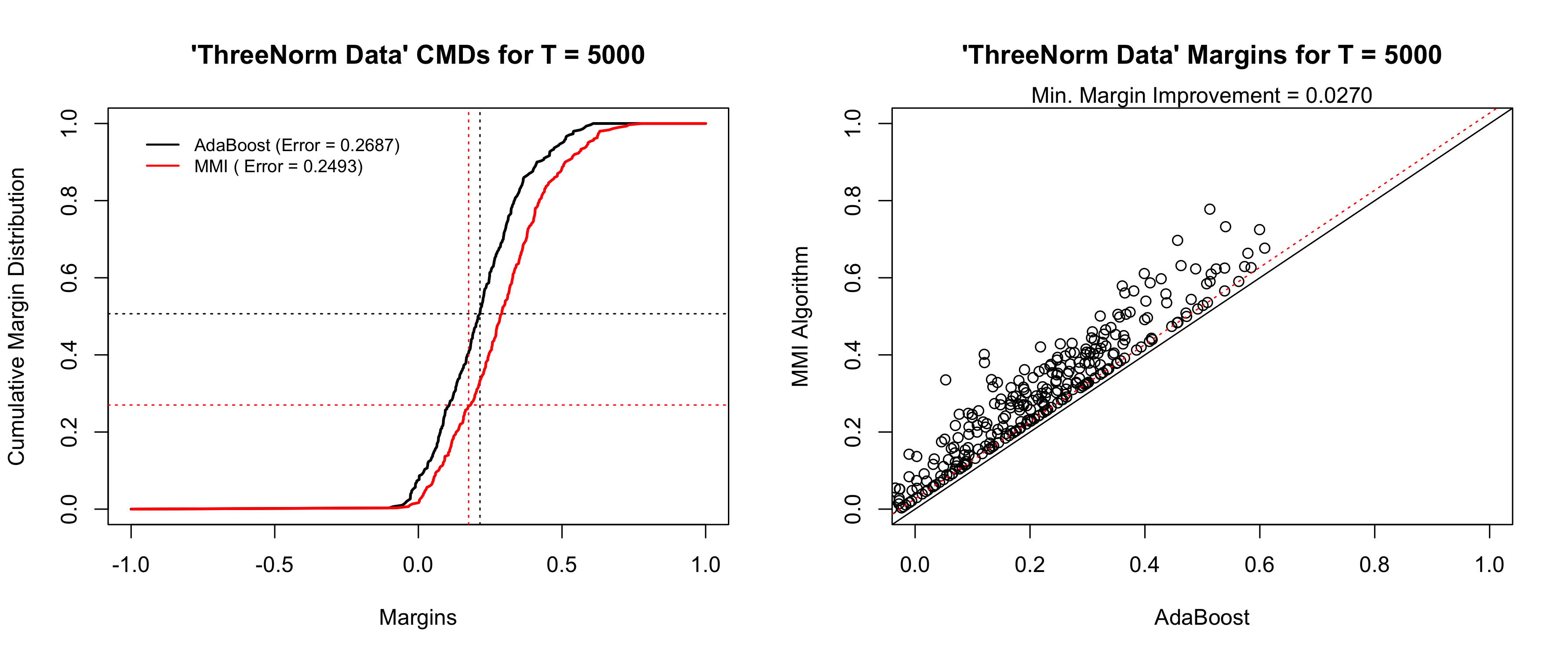}
	\end{minipage}}
 \vfill	
  \subfloat{
	\begin{minipage}[c][1\width]{
	   0.33\textwidth}
	   \centering
	   \includegraphics[width=6.2in, height = 2.4in]{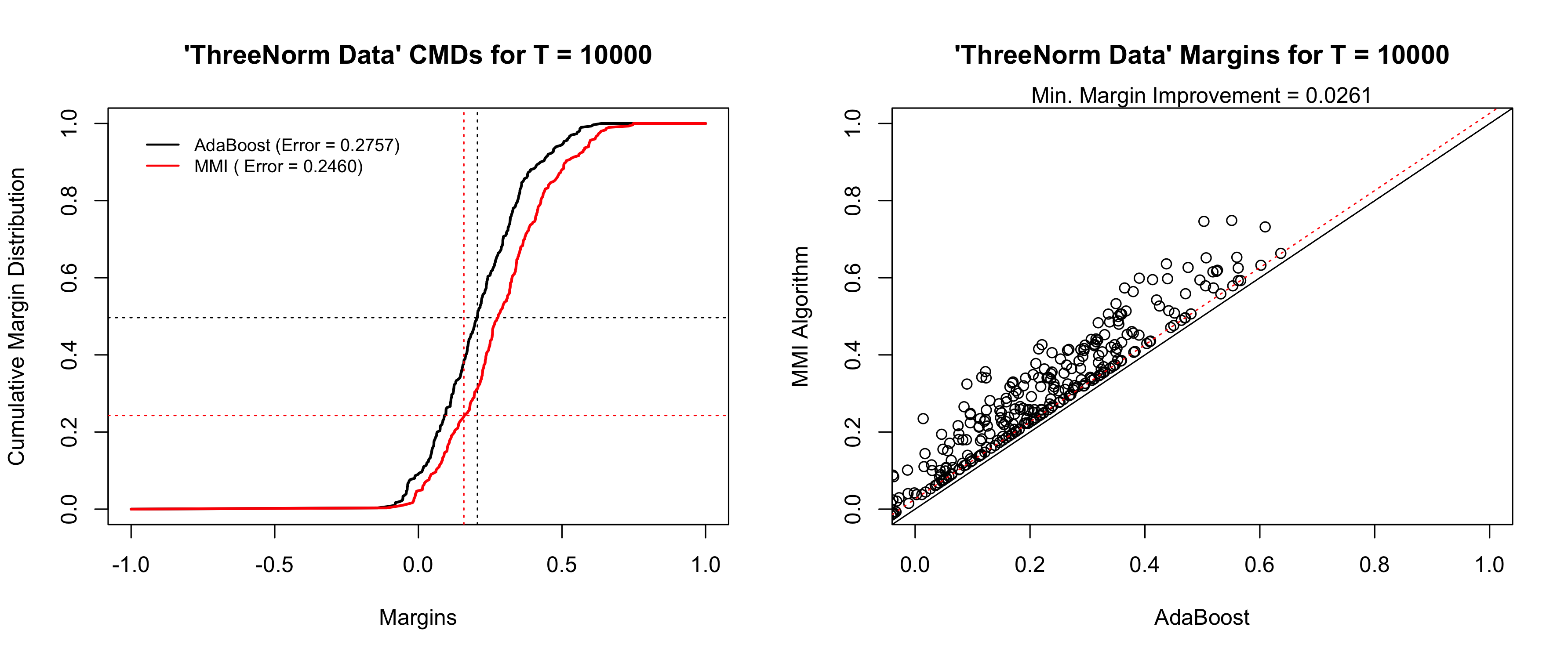}
	\end{minipage}}
\end{center}
\caption{Cumulative margin distributions of AdaBoost versus the MMI Algorithm for the ThreeNorm Data for T = 1000, 5000, 10000 and corresponding margin scatter plots (right) of AdaBoost versus the MMI Algorithm.}
\label{fig6}
\end{figure}

In Figure \ref{fig6}, we visualize the changes to the cumulative margin distributions for high values of $T = 1000, 5000$ and $10000$.  As the number of weak learners combined increases, the minimum margin improvement (and consequently the average margin improvement) generally increases. It is evident for this particular data set that as $T$ increases, there is some amount of overfitting as the test set error degrades. In this particular case, the Emargin continues to increase, while the EMargin error decreases. It is worth mentioning that the bounds in (\ref{eq:5}), (\ref{eq:7}), (\ref{eq:9}) and (\ref{eq:12}) do not depend on $T$.


%

We further test the proposed methods on 20 synthetic and real data sets (see Table \ref{tab:1} for descriptions of data sets). To fix the complexity, we only consider trees of depth = $\{1,2\}$, with $k = \{2,4\}$ respectively. We further normalize each feature to $\left[0,1\right]$ and consider 100 thresholds uniformly distributed on $\left[0,1\right]$ on each feature to be consistent with \cite{wang2011refined}, so that the  $\mathscr{\left|H\right|} = 2 \times 100 \times p$ for decision stumps and $\mathscr{\left|H\right|} = \left(2 \times 100 \times p\right)^3$ for 4-node (depth = 2) trees. The results are summarized in Tables \ref{tab:2}, \ref{tab:3}, \ref{tab:4}, \ref{tab:5}. Tables \ref{tab:2} and \ref{tab:3} illustrate the results for decision stumps using $T = \{250, 500, 750, 1000\}$, while Tables \ref{tab:4} and \ref{tab:5} illustrate the results for 4-node (depth = 2) decision trees also using $T = \{250, 500, 750, 1000\}$.  We use a 70/30 sampling scheme for all data sets, except for those that already have a test set. We show the test set error rate, minimum margin improvement, average margin improvement, the EMargin and EMargin error for all data sets and experiments. Bold EMargins and EMargin errors indicate inconsistencies with the EMargin theory and the bound in (\ref{eq:9}) for Theorem 3. The results from using decision stumps indicate that the MMI algorithm cannot easily find optimized weights which improve upon the margin distribution resulting from the AdaBoost solution. Only when the number of combined classifiers $T$ is 750 and 1000 can we find changes in the margin distributions on specific data sets, such as the ColonCancer, Mushrooms, Musk, Sonar and TwoNorm data sets. For those simulations only the Sonar data set shows an improved performance, while the performance of the TwoNorm data set worsened and the rest of the data sets, even with improved margin distributions, have the same test set error performance.  The results indicate that for 250 and 500 decision stumps, there is no difference in performance in terms of the test set error rate between AdaBoost and the MMI algorithm, and only two out of the 20 data sets had their margins distributions improved by MMI. For 750 and 1000 decision stumps, the AdaBoost algorithm performs better in one data set, while the MMI algorithm performed better in one data set. The rest of the data sets had exactly the same performance. The MMI algorithm was able to improve the margin distributions of six data sets. The greater improvements to the margins of data sets can be found using 4-node (depth = 2) trees. For instance 12 and 14 data sets had their margins distributions improved when $T = 250$ and $T = 500$ respectively. The AdaBoost algorithm performed better in 2 data sets out of 20, while the MMI algoritm performed better also in 2 out of the 20, while the rest were ties. AdaBoost performed better in 3 out of 20, while MMI performed better in 2 out of 20 using 500 decision trees. For 750 decision trees, virtually all data sets had improvements in their margins distributions, with AdaBoost outperforming MMI in 4 data sets, while MMI performed better in just 2. Surprisingly, for 1000 trees, MMI performed better in 6 data sets, while AdaBoost performed better in just 1. 

\begin{table*}[t]
	\centering
	\caption{Description of Data Sets}
	\begin{tabular}{lllccc}
		\toprule
		Data Set & Description &  Source & Training & Testing  & Features  \\
		\midrule
		Australian &  Credit Approval&\cite{Lichman2013} &690 &         & 14      \\
		BreastCancer & Breast Cancer Wisconsin & \cite{Lichman2013}  &683   &       & 10    \\
		ColonCancer & Colon Cancer Data & \cite{alon1999broad}  &62   &       & 2000    \\
		Diabetes & Diabetes Patient Records & \cite{Lichman2013} &768   &       & 8     \\
		FourClass & Non-Separable  &  \cite{Lichman2013} &862   &       & 2     \\
		Gissette & Digit Recognition Data &\cite{guyon2004result}  &6000  & 1000  & 5000   \\	
		IJCNN1 & IJCNN 2001 Competition & \cite{prokhorov2001slide}  &49990 & 91701 & 22   \\
		Ionosphere & Ionosphere Data Set & \cite{Lichman2013}  &351 &  & 34    \\
		Madelon & Artificial Data & \cite{guyon2004result}  &2000  & 600   & 500   \\
		Mushrooms & Mushrooms Data Set& \cite{Lichman2013}  &8124  &    & 112   \\
		Musk & Molecules Prediction& \cite{Lichman2013}  &6598  &    & 168  \\
		Parkinsons & Parkinsons Disease& \cite{Lichman2013}  &197  &    & 23  \\
		Pima& Pima Indians Data Set& \cite{Lichman2013}  &768  &    & 8   \\
		RingNorm & Breiman RingNorm & \cite{Lichman2013} &210 &  90  & 20   \\
		Sonar  & Sonar Data Set  & \cite{Lichman2013} &208  &   & 60    \\
		Spambase& Spam Emails Data Set& \cite{Lichman2013}  &4601  &    & 57   \\
		Splice  & DNA Splice Junctions   & \cite{Lichman2013} &1000  & 2175  & 60   \\	
		Transfusion	& Blood Transfusion & \cite{Lichman2013}  &748 &    & 5   \\
		TwoNorm & Breiman TwoNorm  & \cite{Lichman2013} &300 &  3000  & 20   \\
		ThreeNorm & Breiman ThreeNorm & \cite{Lichman2013} &300 & 3000   & 20   \\
		\bottomrule
	\end{tabular}%
	\label{tab:1}%
\end{table*}%

Figure 7 illustrates visually the changes to the margin distributions obtained by the MMI algorithm for 1000 trees (4-node, depth = 2), in which the MMI algorithm outperformed AdaBoost in most data sets. It is not evident that the improvements to the margin distributions differed from other settings here.

	\begin{table*}[htbp]
		\centering
		\caption{AB vs MMI Using Decision Stumps (MI = margin improvement, EM = EMargin).}
			\resizebox{\textwidth}{!}{%
		\begin{tabular}{llcccccccccc}
			\toprule
			& &\multicolumn{5}{c}{250 Trees ($k=2$, depth = 1)} & \multicolumn{5}{c}{500 Trees ($k=2$, depth = 1)}   \\
			\midrule
			& \multicolumn{1}{l}{\textbf{}} & \multicolumn{1}{c}{Test Error} & \multicolumn{1}{c}{Min. MI} & \multicolumn{1}{c}{Avg. MI} & \multicolumn{1}{c}{EM} & \multicolumn{1}{c}{EM Error} & \multicolumn{1}{c}{Test Error} & \multicolumn{1}{c}{Min. MI} & \multicolumn{1}{c}{Avg. MI} & \multicolumn{1}{c}{EM} & \multicolumn{1}{c}{EM Error} \\
						\midrule
			Australian & AB & 0.1106 & -  & - & 0.3808 & 0.3776    & 0.1154&- & -  & 0.2854 & 0.3714 \\
			& MMI & 0.1106  & 0.0000   & 0.0000& 0.3808 & 0.3776    &  0.1154  & 0.0000 & 0.0000   & 0.2854 & 0.3714 \\
			\midrule
			BreastCancer & AB & 0.0146& -   & -& 0.4040& 0.1925  & 0.0195 & -  & -    & 0.3193& 0.1778\\
					& MMI & 0.0146& 0.0000  & 0.0000 & 0.4040 & 0.1925   & 0.0195 & 0.0000 & 0.0000   & 0.3193 & 0.1778 \\
					\midrule
			ColonCancer & AB & 0.1579 & -    & - & 0.5551 & 0.0232   & 0.1579 & - & -   & 0.5667 & 0.0232\\
					& MMI & 0.1579& 0.0001   & 0.0015 & 0.5551& 0.0232     & 0.1579& 0.0003& 0.0016   & 0.5667& 0.0232 \\
					\midrule
			Diabetes   & AB & 0.2511 & -   & - & \textbf{0.2649} & \textbf{0.5177}    & 0.2467 & - & -  & \textbf{0.2302} & \textbf{0.5233} \\
					& MMI & 0.2511 & 0.0000  & 0.0000 & 0.2649 & 0.5177    & 0.2467 & 0.0000 & 0.0000  & 0.2302& 0.5233 \\
					\midrule
			FourClass & AB & 0.2124 & -   & - & 0.6979    & 0.7546    & 0.2085 & - &-   & 0.5791 & 0.7546 \\
					& MMI & 0.2124 & 0.0000   & 0.0000 & 0.6979 & 0.7546    & 0.2085 & 0.0000& 0.0000   & 0.5791 & 0.7546 \\
					\midrule
			Gissette & AB & 0.0419 & -   &-& 0.2775 & 0.4354    & 0.0419& -& -   & 0.2738 & 0.5225 \\
& MMI & 0.0419 & 0.0000   & 0.0000 & 0.2775& 0.4354    & 0.0419& 0.0000 & 0.0000  & 0.2738 & 0.5225\\
\midrule
			IJCNN1	& AB & 0.1053 &  -   & - & \textbf{0.1862 }& \textbf{0.1401}   & 0.1083 & - & -   & \textbf{0.1577} & \textbf{0.1475}\\
					& MMI & 0.1053  & 0.0000   & 0.0000& 0.1862 & 0.1401    & 0.1083  & 0.0000 & 0.0000   & 0.1577& 0.1475\\
					\midrule	
			Ionosphere & AB &0.0755& -  & - & 0.3180 & 0.2980    & 0.0472 & -  & -    & 0.2140 & 0.2694\\
			& MMI & 0.0755 & 0.0000   & 0.0000 & 0.3180& 0.2980    & 0.0472 & 0.0000 & 0.0000   & 0.2140 & 0.2694 \\
			\midrule
			Madelon & AB &  0.4167 & -   & - & \textbf{0.2050} & \textbf{0.6245}     & 0.4100 & - & -   & \textbf{0.1567} & \textbf{0.6700} \\
					& MMI &  0.4167 & 0.0000   & 0.0000 & 0.2050 & 0.6245     & 0.4100 & 0.0000 & 0.0000  & 0.1567 & 0.6700 \\
					\midrule
			Mushrooms & AB & 0.0172 & -  & - & 0.2702& 0.1081   & 0.0029& - & -  & 0.1709& 0.0568\\
					& MMI & 0.0172 & 0.0000   & 0.0000 & 0.2702 & 0.1081    & 0.0029& 0.0000 & 0.0000  & 0.1709 & 0.0568 \\
					\midrule
			Musk &  AB & 0.2238 & -  & - & \textbf{0.2480} & \textbf{0.6723}   & 0.1888 & -& -   & \textbf{0.2191} & \textbf{0.7117} \\
			& MMI & 0.2238 & 0.0000  & 0.0000& 0.2480& 0.6723    & 0.1888 & 0.0000 & 0.0000   & 0.2191& 0.7117\\
			\midrule
			Parkinsons    & AB & 0.1864& -  & - & 0.3360 & 0.3971    & 0.1525 & - & -   & 0.2576 & 0.3162 \\
					& MMI & 0.1864 & 0.0000  & 0.000& 0.3360  & 0.3971    & 0.1695 & 0.0001& 0.0067   & 0.2659& 0.3529 \\
					\midrule
			Pima & AB & 0.2511 & -   & - & \textbf{0.2649}& \textbf{0.5177}   & 0.2467 & -&-   & \textbf{0.2302} & \textbf{0.5233} \\
					& MMI &0.2511 & 0.0000   & 0.0000 & 0.2649& 0.5177    & 0.2467  & 0.0000& 0.0000  & 0.2302 & 0.5233 \\
					\midrule	
			RingNorm & AB & 0.3180  & -   & - & \textbf{0.1588} & \textbf{0.4667}   & 0.2800& - & -  & \textbf{0.1262} & \textbf{0.4800}\\
					& MMI &0.3180 & 0.0000   & 0.0000 & 0.1588  & 0.4667   & 0.2800  & 0.0000 & 0.0000   & 0.1262& 0.4800\\
					\midrule
			Sonar   & AB & 0.2063 & -  & - & 0.2342 & 0.4000   & 0.2222 & - & -   & 0.2375& 0.4759 \\
					& MMI & 0.2063 & 0.0000   & 0.0000& 0.2342& 0.4000   & 0.2222 & 0.0000 & 0.0000   & 0.2375& 0.4759 \\
					\midrule
			Spambase   & AB & 0.0681 & -  & - & 0.1585& 0.2189  & 0.0673& -& -  & 0.1772 & 0.2807 \\
					& MMI & 0.0681 & 0.0000   &0.0000& 0.1585 & 0.2189    & 0.0673  & 0.0000& 0.0000   &  0.1772 & 0.2807 \\
					\midrule
			Splice   & AB & 0.0818& -  & - & 0.1930 & 0.3240    & 0.0662 & - & -   & 0.1561 & 0.3090 \\
						& MMI & 0.0818 & 0.0000  & 0.0000 & 0.1930 & 0.3240    & 0.0662& 0.0000 & 0.0000  & 01561& 0.3090\\
					\midrule
			Transfusion   & AB & 0.2133& -   & - & 0.3308 & 0.3174   & 0.2133 & -& -   & 0.3917 & 0.3537 \\
					& MMI & 0.2133 & 0.0000  & 0.0000 & 0.3308 & 0.3174     & 0.2133 & 0.0000 & 0.0000   & 0.3917  & 0.3537 \\	
				\midrule
			TwoNorm   & AB & 0.0697& -   & - & 0.2246& 0.2000   & 0.0677 & -& -   & 0.1984 & 0.1667 \\
					& MMI & 0.0697 & 0.0000  & 0.0000 & 0.2246& 0.2000    & 0.0677 & 0.0000 & 0.0000   & 0.1984  & 0.1667\\
		           \midrule
		    ThreeNorm   & AB & 0.2667 & -   & - & \textbf{0.1826} & \textbf{0.5533}   & 0.2473& -& -   & \textbf{0.1650} & \textbf{0.5700}\\
			    & MMI & 0.2667  & 0.0000  & 0.0000 & 0.1826 & 0.5533     & 0.2473  & 0.0000 & 0.0000   & 0.1650  & 0.5700 \\	
			    \midrule
			    Summary & &  & & & & & & & & & \\
			    			    \midrule
			     & AB & 0 wins &    &  & & &0 wins & &    & &  \\
			     &  MMI &0 wins& &&& & 0 wins & & & &  \\
				 &   & 20 ties &    & & &  & 20 ties & &   & &  \\
			\bottomrule
		\end{tabular}%
		\label{tab:2}%
	}
	\end{table*}%
	

\begin{table*}[htbp]
	\centering
	\caption{AB vs MMI Using Decision Stumps (MI = margin improvement, EM = EMargin).}
	\resizebox{\textwidth}{!}{%
		\begin{tabular}{llcccccccccc}
			\toprule
			& &\multicolumn{5}{c}{750 Trees ($k=2$, depth = 1)} & \multicolumn{5}{c}{1000 Trees ($k=2$, depth = 1)}   \\
			\midrule
			& \multicolumn{1}{l}{\textbf{}} & \multicolumn{1}{c}{Test Error} & \multicolumn{1}{c}{Min. MI} & \multicolumn{1}{c}{Avg. MI} & \multicolumn{1}{c}{EM} & \multicolumn{1}{c}{EM Error} & \multicolumn{1}{c}{Test Error} & \multicolumn{1}{c}{Min. MI} & \multicolumn{1}{c}{Avg. MI} & \multicolumn{1}{c}{EM} & \multicolumn{1}{c}{EM Error} \\
			\midrule
			Australian & AB & 0.1154 & -  & - & 0.2383 & 0.3714    & 0.1154&- & -  & 0.2276 & 0.3983 \\
			& MMI & 0.1154 & 0.0000   & 0.0000& 0.2383& 0.3714   &  0.1154  & 0.0000 & 0.0000   & 0.2276 & 0.3983 \\
			\midrule
			BreastCancer & AB & 0.0195& -   & -& 0.3122 & 0.2008   & 0.0195 & -  & -    & 0.2861& 0.2008 \\
			& MMI & 0.0195& 0.0000  & 0.0000 & 0.3122 & 0.2008 & 0.0195 & 0.0000 & 0.0000   & 0.2861 & 0.2008 \\
			\midrule
			ColonCancer & AB & 0.1579 & -    & - & 0.5725 & 0.0233   & 0.1579 & - & -   & 0.5752 & 0.0233\\
			& MMI & 0.1579& 0.0004   & 0.0041 & 0.5725& 0.0233    & 0.1579& 0.0003& 0.0031  & 0.5752& 0.0233 \\
			\midrule
			Diabetes   & AB & 0.2511 & -   & - & 0.2139 & 0.5456   & 0.2597 & - & -  & 0.2041& 0.5568 \\
			& MMI & 0.2511 & 0.0000  & 0.0000 & 0.2139 & 0.5456    & 0.2597 & 0.0000 & 0.0000  & 0.2041& 0.5568 \\
			\midrule
			FourClass & AB & 0.1930& -  &- & 0.5096 & 0.7546   & 0.1930& -  &- & 0.5096 & 0.7546 \\
			& MMI & 0.1930& 0.0000   & 0.0000& 0.5096& 0.7546    & 0.1930& 0.0000   & 0.0000& 0.5096& 0.7546 \\
			\midrule
			Gissette & AB & 0.0419 & -   &-& 0.2775 & 0.4354    & 0.0419& -& -   & 0.2738 & 0.5225 \\
			& MMI & 0.0419 & 0.0000   & 0.0000 & 0.2775& 0.4354    & 0.0419& 0.0000 & 0.0000  & 0.2738 & 0.5225\\
			\midrule
			IJCNN1	& AB & 0.1071& -   & - & 0.1374 & 0.1481   & 0.1079 & - & -   & 0.1343 & 0.1584 \\
			& MMI & 0.1071  & 0.0000   & 0.0000 & 0.1374  & 0.1481   & 0.1079 & 0.0000 & 0.0000  & 0.1343 & 0.1584 \\
			\midrule	
			Ionosphere & AB & 0.0566 & -   & - & 0.1728 & 0.2490    & 0.0566& - & -    & 0.1847 & 0.3551 \\
			& MMI & 0.0566 & 0.0000  & 0.0000 & 0.1728& 0.2490    & 0.0566 & 0.0000  & 0.0000   & 0.1847 & 0.3551 \\
			\midrule  
			Madelon & AB & 0.2238 & -  & - & \textbf{0.2480} & \textbf{0.6723}   & 0.1888 & -& -   & \textbf{0.2191}& \textbf{0.7117} \\
			& MMI & 0.2238 & 0.0000  & 0.0000& 0.2480& 0.6723    & 0.1888 & 0.0000 & 0.0000   & 0.2191& 0.7117\\
			\midrule
			Mushrooms & AB & 0.0028 &-   &-& 0.1735& 0.0556    & 0.0016 & -  & -   & 0.1673 & 0.0556\\
			& MMI & 0.0028 & 0.0000   & 0.0072& 0.1735& 0.0598  & 0.0016& 0.0013 & 0.0062  & 0.1578 & 0.0492 \\
			\midrule
			Musk & AB & 0.1538& -   & - & 0.1874 & 0.7087   & 0.1538 & -& -   & 0.1604 & 0.6396 \\
			& MMI & 0.1538 & 0.0000   & 0.0299 & 0.1874& 0.7087    & 0.1538 & 0.0000& 0.0315   & 0.2673 & 0.9429\\
			\midrule
			Parkinsons    & AB & 0.1525 & -    & - & 0.2762& 0.4191   & 0.1525 & - & -  & 0.2439 & 0.3823 \\
			& MMI & 0.1525 & 0.0006   & 0.0086 & 0.2595& 0.3971    & 0.1525 & 0.0013 & 0.0089   & 0.2460 & 0.3971\\
			\midrule
			Pima & AB & 0.2467 & -    & - & 0.2302 & 0.5233    & 0.2597 & -  & -   & 0.2040 & 0.5568\\
			& MMI &0.2467 & 0.0000   & 0.0000 & 0.2302 & 0.5233    & 0.2597 & 0.0000 & 0.0000   & 0.2040 & 0.5568 \\
			\midrule	
			RingNorm & AB & 0.2603 & -   & - & 0.1188 & 0.4967  & 0.2433& - & -   & 0.1003 & 0.4033\\
			& MMI &0.2603 & 0.0000   & 0.0000 & 0.1188 & 0.4967     & 0.2433  & 0.0000 & 0.0000  & 0.1003  & 0.4033 \\
			\midrule
			Sonar   & AB & 0.2222& -  & - & 0.2187& 0.4759  & 0.2063 & - & -   & 0.2029& 0.4759 \\
			& MMI & 0.1905 & 0.0010  & 0.0062& 0.2103& 0.4690   & 0.2063 & 0.0021 & 0.0114   & 0.1895& 0.4483 \\
			\midrule
			Spambase   & AB & 0.0667 & -   & -& 0.1616 & 0.2779   & 0.0637 & - & -   & 0.1455 & 0.2689 \\
			& MMI & 0.0667 & 0.0000  & 0.0000& 0.1616& 0.2779   & 0.0637 & 0.0000 & 0.0000   & 0.1455 & 0.2689\\
			\midrule
			Splice   & AB & 0.0621& -  & - & 0.1685 & 0.3820   & 0.0607& - & -   & 0.1763& 0.4310 \\
			& MMI & 0.0621 & 0.0000  & 0.0000 & 0.1685 & 0.3820    & 0.0607& 0.0000 & 0.0000  & 0.1763& 0.4310\\
			\midrule
			Transfusion   & AB & 0.2133& -   & - & 0.3803 & 0.3384   & 0.2133 & -& -   & 0.3571& 0.3155 \\
			& MMI & 0.2133 & 0.0000  & 0.0000 & 0.3803 & 0.3384    & 0.2133 & 0.0000 & 0.0000   & 0.3571 & 0.3155\\	
				\midrule
			TwoNorm   & AB & 0.0647& -   & - & 0.1923 & 0.1600  & 0.0650 & -& -   & 0.1874 & 0.1633 \\
			& MMI & 0.0660& 0.0000  & 0.0042 & 0.1939 & 0.1667    & 0.0653 & 0.0023 & 0.0100   & 0.2007  & 0.2100 \\
			\midrule
			ThreeNorm   & AB & 0.2350 & -   & - & 0.1442  & 0.5100   & 0.2283 & -& -   & 0.1472 & 0.5600  \\
			& MMI & 0.2350 & 0.0000  & 0.0000 & 0.1442 & 0.5100    & 0.2283 & 0.0000 & 0.0000   & 0.1472 & 0.5600 \\	
						    \midrule
			Summary & &  & & & & & & & & & \\
			\midrule
			& AB & 1 wins &    &  & & &1 wins & &    & &  \\
			&  MMI &1 wins& &&& & 1 wins & & & &  \\
			&   & 18 ties &    & & &  & 18 ties & &   & &  \\
			\bottomrule
		\end{tabular}%
		\label{tab:3}%
	}
\end{table*}%

\begin{table*}[htbp]
	\centering
	\caption{AB vs MMI Using 4-node (depth = 2) Trees (MI = margin improvement, EM = EMargin)}
	\resizebox{\textwidth}{!}{%
		\begin{tabular}{llcccccccccc}
			\toprule
			& &\multicolumn{5}{c}{250 Trees ($k=4$, depth = 2)} & \multicolumn{5}{c}{500 Trees ($k=4$, depth = 2)}   \\
			\midrule
			& \multicolumn{1}{l}{\textbf{}} & \multicolumn{1}{c}{Test Error} & \multicolumn{1}{c}{Min. MI} & \multicolumn{1}{c}{Avg. MI} & \multicolumn{1}{c}{EM} & \multicolumn{1}{c}{EM Error} & \multicolumn{1}{c}{Test Error} & \multicolumn{1}{c}{Min. MI} & \multicolumn{1}{c}{Avg. MI} & \multicolumn{1}{c}{EM} & \multicolumn{1}{c}{EM Error} \\
			\midrule
			Australian& AB & 0.1202 & -  & - & 0.3151  & 0.3008 & 0.1202 & - & -   &  0.2740  & 0.3361\\
			& MMI & 0.1202  & 0.0000   & 0.0026 & 0.3116 & 0.2987    & 0.1202& 0.0000 & 0.0025   & 0.2740  & 0.3361 \\
			\midrule
			BreastCancer & AB & 0.0098& -   & - & 0.3822 & 0.1611  & 0.0098 & - &-    & 0.3147 & 0.1464 \\
			& MMI & 0.0098 & 0.0027 & 0.0198 & 0.3738 & 0.1569    &0.0098 & 0.0095 & 0.0380   & 0.2894 & 0.1297 \\
			\midrule
			ColonCancer & AB & 0.1579 & -    & - & 0.5725 & 0.0233   & 0.1579 & - & -   & 0.5752 & 0.0233\\
			& MMI & 0.1579& 0.0004   & 0.0041 & 0.5725& 0.0233    & 0.1579& 0.0003& 0.0031  & 0.5752& 0.0233 \\
			\midrule
			Diabetes   & AB & 0.2424 & -  & -& 0.2793 & 0.4693    & 0.2424 & - & -  & 0.2451 & 0.5065 \\
			& MMI & 0.2424 & 0.0000   & 0.0004 & 0.2793& 0.4693   & 0.2424 & 0.0000 & 0.0000   & 0.2451 & 0.5065 \\
			\midrule
			FourClass & AB & 0.0656 & -   &- & 0.2342 & 0.4461  & 0.0386 & - & -   & 0.1805 & 0.3482 \\
			& MMI & 0.0656& 0.0000  & 0.0013& 0.2342& 0.4461    & 0.0386 & 0.0097 & 0.0312   & 02091& 0.4146 \\
			\midrule
			Gissette & AB & 0.0399 & -   &-& 0.2365 & 0.4356    & 0.0399& -& -   & 0.2738 & 0.5225 \\
			& MMI & 0.0399 & 0.0000   & 0.0000 & 0.2365& 0.4356    & 0.0399& 0.0000 & 0.0000  & 0.2778 & 0.5225\\
			\midrule
			IJCNN1	& AB & 0.1010 & -   & - & \textbf{0.1481} & \textbf{0.1071}   & 0.1005 & -  & -    & \textbf{0.1420} & \textbf{0.1183} \\
		    & MMI & 0.1009 & 0.0000   & 0.0000 & 0.1482 & 0.1071   & 0.1005 & 0.0000 & 0.0000   & 0.1420& 0.1183\\
			\midrule
			Ionosphere & AB & 0.0377 & -   & -  & 0.2849 & 0.1510  & 0.0566& - & -    & 0.2546 & 0.1102 \\
			& MMI & 0.0377 & 0.0031 & 0.0127 & 0.2849 & 0.1510   & 0.0566& 0.0184 & 0.0390  & 0.2373& 0.0939\\
			\midrule
		Madelon & AB & 0.2238 & -  & - & \textbf{0.2480} & \textbf{0.6723}   & 0.1888 & -& -   & \textbf{0.2191}& \textbf{0.7117} \\
		& MMI & 0.2238 & 0.0000  & 0.0000& 0.2480& 0.6723    & 0.1888 & 0.0000 & 0.0000   & 0.2191& 0.7117\\
		\midrule
			Mushrooms & AB & 0.0000 & -   & - & 0.2464 & 0.0227    & 0.0000 & -& -   & 0.2454& 0.0260\\
			& MMI & 0.0000 & 0.0323  & 0.0617& 0.1878& 0.0188   & 0.0000& 0.1601 & 0.1799  & 0.1604 & 0.0095\\
			\midrule
			Musk & AB & 0.1399 & -   &-& 0.2365 & 0.4354    & 0.1818& -& -   & 0.2738 & 0.5225 \\
			& MMI & 0.1399 & 0.0000   & 0.0002 & 0.2365& 0.4354    & 0.1748& 0.0022 & 0.0145   & 0.2778 & 0.5345\\
			\midrule
			Parkinsons    & AB & 0.0678 & -   & - & 0.3562 & 0.1250    & 0.0847 &- & -   & 0.2855& 0.0294 \\
			& MMI & 0.0508& 0.0558  & 0.1060 & 0.12803 & 0.0367    & 0.0678& 0.0657 & 0.1238   & 0.2855 & 0.0294 \\
			\midrule
			Pima & AB & 0.2424 & -   & - & 0.2793 & 0.4693    & 0.2424& -& -  & 0.2451 & 0.5065 \\
			& MMI &0.2424 & 0.0000   & 0.0004 & 0.2793 & 0.4693    & 0.2424 & 0.0000& 0.0000   & 0.2451 & 0.5065 \\
			\midrule
			RingNorm & AB & 0.2090 & -   & -  & 0.1949 & 0.3233   & 0.1533 & - & -    & 0.1528 & 0.2967 \\
			& MMI &0.2090   & 0.0000   & 0.0000 & 0.1949  & 0.3233    & 0.1533 & 0.0000 & 0.0001  & 0.1528 & 0.2967 \\
			\midrule
			Sonar   & AB & 0.1746 & -  & - & 0.2659& 0.0345    & 0.2063 & - & -  & 0.2502& 0.0207\\
			& MMI & 0.2063 & 0.0159   & 0.0359 & 0.2535 & 0.0275    & 0.2222 & 0.0206& 0.0415   & 0.2390& 0.0138 \\
			\midrule
			Spambase   & AB & 0.0637 & - & - & \textbf{0.1455} & \textbf{0.2689}    & 0.0659& -  & -    & \textbf{0.1916} & \textbf{0.2208}\\
			& MMI & 0.0637 & 0.0000  & 0.0000 & 0.1455 & 0.2689    & 0.0659& 0.0000 & 0.0001   & 0.1916 & 0.2208 \\
			\midrule
			Splice   & AB & 0.0336 & - & -   & 0.2301 & 0.2260  & 0.0349 & -& -  & 0.1899 & 0.2160 \\
			& MMI & 0.0336 & 0.0000& 0.0000 & 0.2301 & 0.2260 & 0.0349  & 0.0000 & 0.0000   & 0.1899 & 0.2160\\
			\midrule
			Transfusion   & AB & 0.2133 & -   & - & \textbf{0.3171} &\textbf{0.3575}    & 0.2000 & - & -   & \textbf{0.2590} & \textbf{0.3632} \\
			& MMI & 0.2133& 0.0000  & 0.0000 & 0.3171 & 0.3575    & 0.2000& 0.0000 & 0.0000   & 0.2590 & 0.3632 \\
				\midrule
			TwoNorm   & AB & 0.0747& -   & - & 0.3078 & 0.0167  & 0.0613& -& -   & 0.2871 & 0.0133 \\
			& MMI & 0.0827 & 0.0113  & 0.0287 & 0.2933 & 0.0133     & 0.0707  & 0.0244 & 0.0458    & 0.2871  & 0.0133\\
			\midrule
			ThreeNorm   & AB & 0.2093& -   & - & \textbf{0.1982} & \textbf{0.2300}   & 0.2030& -& -   & \textbf{0.1878} & \textbf{0.2400} \\
			& MMI & 0.2063 & 0.0067  & 0.0000 & 0.2029 & 0.2533    & 0.2063 & 0.0153 & 0.0419  & 0.1797 & 0.2200 \\	
						    \midrule
			Summary & &  & & & & & & & & & \\
			\midrule
			& AB & 2 wins &    &  & & &3 wins & &    & &  \\
			&  MMI &2 wins& &&& & 2 wins & & & &  \\
			&   & 16 ties &    & & &  & 15 ties & &   & &  \\
			\bottomrule
		\end{tabular}%
		\label{tab:4}%
	}
\end{table*}%

\begin{table*}[htbp]
	\centering
	\caption{AB vs MMI Using 4-node (depth = 2) Trees (MI = margin improvement, EM = EMargin)}
	\resizebox{\textwidth}{!}{%
		\begin{tabular}{llcccccccccc}
			\toprule
			& &\multicolumn{5}{c}{750 Trees ($k=4$, depth = 2)} & \multicolumn{5}{c}{1000 Trees ($k=4$, depth = 2)}   \\
			\midrule
			& \multicolumn{1}{l}{\textbf{}} & \multicolumn{1}{c}{Test Error} & \multicolumn{1}{c}{Min. MI} & \multicolumn{1}{c}{Avg. MI} & \multicolumn{1}{c}{EM} & \multicolumn{1}{c}{EM Error} & \multicolumn{1}{c}{Test Error} & \multicolumn{1}{c}{Min. MI} & \multicolumn{1}{c}{Avg. MI} & \multicolumn{1}{c}{EM} & \multicolumn{1}{c}{EM Error} \\
			\midrule
			Australian& AB & 0.1250 & -  & - & 0.2610 & 0.3880& 0.1250 & - & -   &  0.2105 & 0.3589\\
			& MMI & 0.1250  & 0.0000   & 0.0021 & 0.2590 & 0.3860   & 0.1250& 0.0000 & 0.0032  & 0.2053 & 0.3527\\
			\midrule
			BreastCancer & AB & 0.0098& -   & - & 0.2902& 0.1485  & 0.0098 & - &-    & 0.2912 & 0.1653 \\
			& MMI & 0.0098 & 0.0126 & 0.0505& 0.2778 & 0.1443   &0.0098 & 0.0145 & 0.0560   & 0.2678& 0.1506 \\
			\midrule
			ColonCancer & AB & 0.1579 & -    & - & 0.5725 & 0.0233   & 0.1579 & - & -   & 0.5752 & 0.0233\\
			& MMI & 0.1579& 0.0004   & 0.0041 & 0.5725& 0.0233    & 0.1579& 0.0003& 0.0031  & 0.5752& 0.0233 \\
			\midrule
			Diabetes   & AB & 0.2554 & -  & -& \textbf{0.2431} & \textbf{0.5754}    & 0.2424 & - & -  & \textbf{0.2271} & \textbf{0.5978} \\
			& MMI & 0.2554& 0.0000   & 0.0004 & 0.2431& 0.4754   & 0.2424 & 0.0000 & 0.0000   & 0.2271 & 0.5978 \\
			\midrule
			FourClass & AB & 0.0154& -    & -& 0.1651 & 0.2935    & 0.0154& - & -   & 0.1426 & 0.1990\\
			& MMI & 0.0154 & 0.0126  & 0.0348& 0.1690& 0.3184   & 0.0116& 0.0125 & 0.0328   & 0.1273 & 0.1426 \\
			\midrule
			Gissette & AB & 0.0399 & -   &-& 0.2365 & 0.4354    & 0.0218& -& -   & 0.2738 & 0.5225 \\
			& MMI & 0.0399 & 0.0000   & 0.0000 & 0.2365& 0.4354    & 0.0218& 0.0000 & 0.0000  & 0.2738 & 0.5225\\
			\midrule
			IJCNN1	& AB & 0.0999 & -   & -& 0.1370 & 0.1229    & 0.1013 & - & -   & 0.1259 & 0.1217\\
			& MMI & 0.1000 & 0.0000   & 0.0001 & 0.1370 & 0.1229    & 0.1013  & 0.0000 & 0.0000   & 0.1259 & 0.1217 \\
			
			\midrule
			Ionosphere & AB & 0.0660& -  & - & 0.2164 & 0.0612    & 0.0660& -  & -    & 0.2137 & 0.0531\\
			& MMI & 0.0600 & 0.0239  & 0.0526 & 0.2033 & 0.0449   & 0.0566 & 0.0292 & 0.0573   & 0.1943 & 0.0367 \\
			\midrule
			Madelon & AB & 0.4167 & -   & - & 0.2050 & 0.6245    & 0.3550  & - & -   & 0.1252 & 0.6870\\
			& MMI & 0.4167 & 0.0000   & 0.0000 & 0.2050 & 0.6245    & 0.3600 & 0.0000 & 0.0001   & 0.1246 & 0.6860 \\
			\midrule
			Mushrooms & AB & 0.0000 & -   & - & 0.2385& 0.0237    & 0.0000& - & -   & 0.1615& 0.0058 \\
			& MMI & 0.0000& 0.0348   & 0.0768 & 0.1838 & 0.0128    & 0.0000 & 0.0351 & 0.0738   & 0.1646 & 0.0100 \\
			\midrule
			Musk & AB & 0.2308 & -  & - & 0.2536 & 0.4715   & 0.2797& -& -   & 0.2723& 0.4835 \\
			& MMI & 0.2308 & 0.0038 & 0.0149& 0.2658& 0.5015   & 0.2657& 0.0048 & 0.0196   & 0.2835& 0.5105\\
			\midrule
			Parkinsons    & AB & 0.0847 & -  &- & 0.2902& 0.0368    & 0.0847& -& -   & 0.3439& 0.1323\\
			& MMI & 0.0847 & 0.0639   & 0.1285 & 0.2902 & 0.0368    & 0.0847 & 0.0628 & 0.1270  & 0.2651 & 0.0294\\
			\midrule
			Pima & AB & 0.2554& -   & - & \textbf{0.2431}& \textbf{0.5754}    & 0.2424& - & -   & \textbf{0.2271}& \textbf{0.5978} \\
			& MMI &0.2554 & 0.0000   & 0.0000 & 0.2431 & 0.5754    & 0.2424 & 0.0000 & 0.0000   & 0.2271 & 0.5978 \\
			\midrule
			RingNorm & AB & 0.1260& -   & - & 0.1494 & 0.3633  & 0.1160 & - & -   & 0.1323& 0.3333\\
			& MMI &0.1267& 0.0000  & 0.0001 & 0.1494 & 0.3633    & 0.1160 & 0.0002 & 0.0030   & 0.1323& 0.3633\\
			\midrule
			Sonar   & AB & 0.2063 & -  & - & 0.2689& 0.1310    & 0.2063 & - & -  & 0.2948& 0.1241\\
			& MMI & 0.1746 & 0.0214& 0.0496& 0.2581 & 0.0827    & 0.1746& 0.0213& 0.0522   & 0.2498& 0.0897 \\
			\midrule
			Spambase   & AB & 0.0637& -  &- & 0.1763 & 0.2317   & 0.0637& -& -   & 0.1570& 0.2295 \\
			& MMI & 0.0637 & 0.0000   & 0.0001 & 0.1757 & 0.2311   & 0.0637 & 0.0000 & 0.0000  & 0.1564  & 0.2292 \\
			\midrule
			Splice   & AB & 0.0377& - & -   & 0.1983& 0.2620 & 0.0349 & -& -  & 0.1853 & 0.2550 \\
			& MMI & 0.0377 & 0.0000& 0.0000 & 0.1983& 0.2620 & 0.0340  & 0.0002 & 0.0052   & 0.1701 & 0.2240\\
			\midrule
			Transfusion   & AB & 0.2000 & -   & - & 0.2177 &0.3671   & 0.1996& - & -   & 0.3000& 0.5774\\
			& MMI & 0.2100 & 0.0000  & 0.0000 & 0.2177 & 0.3671   & 0.1996& 0.0000 & 0.0000   & 0.3000& 0.5774 \\
				\midrule
			TwoNorm   & AB & 0.0613& -   & - & 0.3131 & 0.0400  & 0.0590 & -& -   & 0.2845 & 0.0200 \\
			& MMI & 0.0673 & 0.0276  & 0.04811 & 0.2776 & 0.0133    & 0.0660 & 0.0276  & 0.0506    & 0.2845  & 0.0200 \\
			\midrule
			ThreeNorm   & AB & 0.2057 & -   & - & 0.1984  & 0.3267   & 0.2093& -& -   & 0.1867 & 0.3133 \\
			& MMI & 0.2050 & 0.0204  & 0.0589 & 0.1775 & 0.2633    & 0.2070 & 0.0231 & 0.0612  & 0.1784  & 0.2900 \\	
						    \midrule
			Summary & &  & & & & & & & & & \\
			\midrule
			& AB & 4 wins &    &  & & & 1 wins & &    & &  \\
			&  MMI &2 wins& &&& & 6 wins & & & &  \\
			&   & 14 ties &    & & &  & 13 ties & &   & &  \\
			\bottomrule
		\end{tabular}%
		\label{tab:5}%
	}
\end{table*}%

\begin{figure*} 
	 \includegraphics[width=\textwidth]{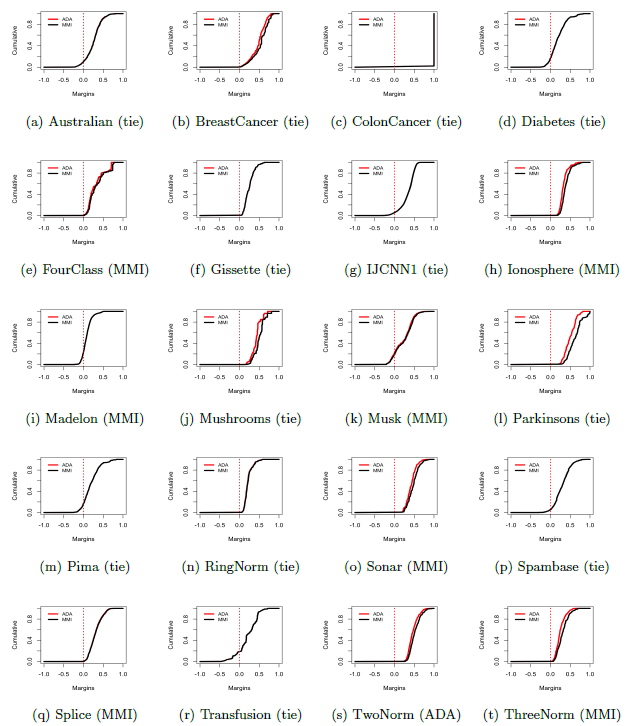}
	\caption{Cumulative Margin Distributions for AdaBoost vs the MMI algorithm using 1000 decision trees ($k=4$, depth = 2). In parenthesis, we can find which algorithm performed better.}
	\label{fig:6}
\end{figure*}

\section{Conclusions}
\label{conc}
\indent \indent In the boosting literature, the upper bound on the generalization error in (\ref{eq:3}) developed by \cite{schapire98} led to the large margins theory, which suggests that, everything else being equal, larger margins should lead to better generalization error (see \citealt{schapire98, grove98, mason2000improved, wang2011refined, shen2010boosting, wang2011refined, wang2012further, gao2013doubt, martinez2014role, liu2015boosting, zhang2016optimal}.) While the upper bound is theoretically correct, the large margins theory appears to be an over-interpretation of that result. The evidence presented in this paper shows that even if all of the margins are increased for a fixed set of weak learners, there is not necessarily an improvement in ensemble performance as measured by test set error, and in some cases the resulting ensemble performs worse. 
As \cite{breiman1999prediction},  \cite{schapire98}  and \cite{wang2011refined} have pointed out, the upper bounds presented in (\ref{eq:3}) and (\ref{eq:4}) are too loose to be practical, so that factors other than the margins may play a major role in generalization error. 
Based on the large margins theory, several researchers have attempted to directly optimize functions of the margins distribution ~\citep{grove98, breiman1999prediction, mason2000improved, ratsch2002maximizing, shen2010boosting, zhou2014large} and have met with mixed success in improving ensemble performance, but none of these algorithms is designed or guaranteed to increase or maintain all of the margins as the MMI algorithm does. This suggests that there may be special functions of margins, or even factors other than margins, that affect generalization error. We have shown that simply improving all of the margins for a fixed set of weak learners is not sufficient to improve generalization error.
Although we do not rule out the importance of margins, we believe that there is room for other factors to influence the performance of boosting and other ensemble methods, and we conclude that the large margins theory, as currently stated, is insufficient in explaining the performance of ensemble methods.



\end{document}